\documentclass[10pt,twocolumn,letterpaper]{article}

\usepackage{cvpr}
\usepackage{times}
\usepackage{epsfig}
\usepackage{graphicx}
\usepackage{amsmath}
\usepackage{enumitem}
\usepackage{amssymb}
\usepackage[ruled]{algorithm2e}
\usepackage{times,amsfonts,color,graphicx,caption,subcaption}

\usepackage[pagebackref=true,breaklinks=true,letterpaper=true,colorlinks,bookmarks=false]{hyperref}

\newcommand{\argmin}{\operatornamewithlimits{argmin}}
\cvprfinalcopy 

\def\cvprPaperID{***} 

\ifcvprfinal\fi
\begin{document}

\title{Online Supervised Hashing for Ever-Growing Datasets}

\author{Fatih Cakir~~~Sarah Adel Bargal~~~Stan Sclaroff\\
Boston University\\
Boston, MA 02215\\
{\tt\small \{fcakir, sbargal, sclaroff\}@cs.bu.edu}
}

\maketitle

\begin{abstract}
Supervised hashing methods are widely-used for nearest neighbor search in computer vision applications. Most state-of-the-art supervised hashing approaches employ batch-learners. Unfortunately, batch-learning strategies can be inefficient when confronted with large training datasets. Moreover, with batch-learners, it is unclear how to adapt the hash functions as a dataset continues to grow and diversify
over time. Yet, in many practical scenarios the dataset grows and diversifies; thus, both the hash functions and the indexing must swiftly accommodate these changes. To address these issues, we propose an online hashing method that is amenable to changes and expansions of the datasets. Since it is an online algorithm, our approach offers linear complexity with the dataset size. Our solution is supervised, in that we incorporate available label information to preserve the semantic neighborhood.  Such an adaptive hashing method is attractive; but it requires recomputing the hash table as the hash functions are updated. If the frequency of update is high, then recomputing the hash table entries may cause inefficiencies in the system, especially for large indexes. Thus, we also propose a framework to reduce hash table updates. We compare our method to state-of-the-art solutions on two benchmarks and demonstrate significant improvements over previous work.
\end{abstract}

\vspace*{-1em}
\section{Introduction}
\label{introduction}
Nearest neighbor search lies at the heart of many computer vision applications. Given a query $\mathbf{x}_q$ and a reference set $\mathcal{X} = \{\mathbf{x}_i\}_{i=1}^N$, the problem is to find $\mathbf{x}^*$ such that $\mathbf{x}^* = {\argmin}_{\mathbf{x} \in \mathcal{X}} d(\mathbf{x},\mathbf{x}_q)$ where $d$ is a particular distance. There are two aspects to the problem: efficient search strategies given the set $\mathcal{X}$ and efficient distance computations. 
The advent of large-scale datasets and the usage of high-dimensional representations have dramatically increased the time and space complexity of search and distance computations. This has inspired researchers to develop techniques that alleviate the ever-mounting computational burden, \eg, 
tree-based construction algorithms \cite{tree1, tree2} and dimensionality reduction methods \cite{isomap,lle}. 

These approaches either do not perform well with high dimensions or do not scale well to large datasets. 
A widely-used approach that does not have the aforementioned drawbacks is \textit{hashing}, in which the memory footprint of the data (or related structures) is significantly reduced, the distance computation is extremely efficient, and  sub-linear search mechanisms are available. 

Many hashing approaches have been proposed 
that preserve similarities as induced by particular metrics and/or available semantic information. However, these methods are generally batch learners: the learning phase takes considerable time and the computations usually grow quadratically with data input size. It is not clear how to adapt the hash codes when a new set of data items arrives. Most procedures may even necessitate learning from scratch. As a result, they are unable to swiftly adapt to changes in datasets. Yet, variations and expansions in datasets are common; making techniques that scale well with the input and are amenable to such variations critical. 

In this paper, we propose an online supervised hashing method that maintains the semantic neighborhood and satisfies the aforementioned properties. Retrieving semantically similar items has its use in many vision applications, including image super-resolution \cite{SuperRes}, scene parsing \cite{LabelTransferCeLiu}, image auto-annotation \cite{tagprop} and label-based image retrieval \cite{Carneiro07supervisedlearning}. Our hashing scheme utilizes Error Correcting Output Codes (ECOCs). ECOCs have their origins in coding theory and have been successfully used to solve many vision related problems \cite{FaceEcoc,ImageEcoc, SparseEcoc}. In this work, ECOCs allow compensation of hash function errors, and thus, increase the robustness of the hash mapping. We also make no prior assumptions on the label space; this is crucial given the ever-growing nature of datasets. Our method allows us to swiftly update the hash functions with incoming data; this makes it adaptive to data variations and diversifications.

An adaptive hashing method is attractive, although it has its own challenges. In particular, the binary codes must be recomputed when the hash mappings are updated. This may cause inefficiencies in the system, especially if such updates are frequent. Therefore, a framework is needed to reduce revisiting binary codes that are already stored in an index.

\noindent In summary, our contributions are twofold: 
 \vspace*{-0.15em}
\begin{enumerate}[label=(\roman*)]
 \setlength\itemsep{0.25em}
\item
We propose an online supervised hashing formulation that directly minimizes an upper bound on the Hamming loss between ECOC hash codes and the outputs of the hash mapping. The ECOC hash codes correspond with labels in the dataset.

\item
We introduce a flexible framework that allows us to reduce hash table entry updates. This is critical, especially when frequent updates may occur as the hash table grows larger and larger.
\end{enumerate}

We learn hash functions orders-of-magnitude faster \vs state-of-the-art batch-learning solutions while achieving comparable performance. We also show improved results over two recent online hashing methods. In addition, we demonstrate better or comparable performance to recent deep learning based hashing approaches without demanding GPU hours for training or fine-tuning a deep network.

\section{Related Work}
\label{related_work}
\subsection{Hashing}
Two distinct hashing approaches exist. The first approach considers mapping similar data points into identical hash bins. A hash lookup strategy is then utilized for a query, returning candidate nearest neighbors from the hash bin that the query is mapped to. Hence, a fraction of the reference set is explored. The second approach is a dimensionality reduction technique with the additional constraint that the mapped representations reside in a Hamming space. With these new \textit{binary} representations, distances can be efficiently computed, allowing even an exhaustive search in the reference set to be done very fast. 

The first group of approaches can be broadly categorized as Locality Sensitive Hashing (LSH) methods \cite{LSH, LSH1, KLSH}. LSH methods have attractive properties: bounds on the query time, low space complexity of related search structures, and the ``goodness'' of the returned nearest neighbor. However, these methods ignore the data distribution and are developed only for certain distance functions. 

For certain applications, distances for a set of data points (training data) are only provided. The particular metric on which these distances are derived may be unknown; therefore, it is necessary to learn hash mappings directly from data. This can be considered as a
Hamming embedding problem, where binary codes are produced for each data point while preserving similarities in the original input space. Such an approach is much more flexible than LSH in that arbitrary similarities can be preserved. 

We can broadly categorize these methods as: spectral techniques \cite{LDAHASH,SH,STH}, order preserving methods \cite{MLH,SSC}, similarity alignment solutions that maximize the equivalence between the input and Hamming space distances \cite{BRE,FastHash,SHK}, quantization based approaches \cite{ITQ,KmeansHashing,PQ} and deep learning based methods \cite{Lin_2015_CVPR_Workshops,LabelTransferCeLiu,aaai_hash}. Recently, supervised approaches have been introduced that preserve the semantic neighborhood \cite{FastHash,SSH}. These methods outperform their unsupervised counterparts when the task is to retrieve semantically similar neighbors. 

Our method is based on Error Correcting Output Codes (ECOCs). ECOCs have  been extensively used in many vision problems. The central idea is the use of codewords with error-correcting abilities. Such error-correctiveness has been exploited in recent supervised hashing methods where a number of hash function errors can be recovered in a mapping. In \cite{ECOH}, a binary ECOC matrix is constructed in which a codeword is considered as a target hash code for a particular label. The codewords constitute the rows of the matrix and the columns denote bipartitions of the data on which the dichotomizers are trained. The dichotomizers are consequently used as hash functions. In \cite{ECC}, the authors propose constructing the ECOC matrix adaptively and learning the hash functions in a Boosting framework to reduce correlations among the hash mappings. 
Using this approach they report state-of-the-art performance for supervised hashing. However both ECOC based methods employ batch-learning; as a result, training takes considerable time even for simple hash functions.

\begin{figure*}[t]
  \centering
      \includegraphics[width=0.75\textwidth]{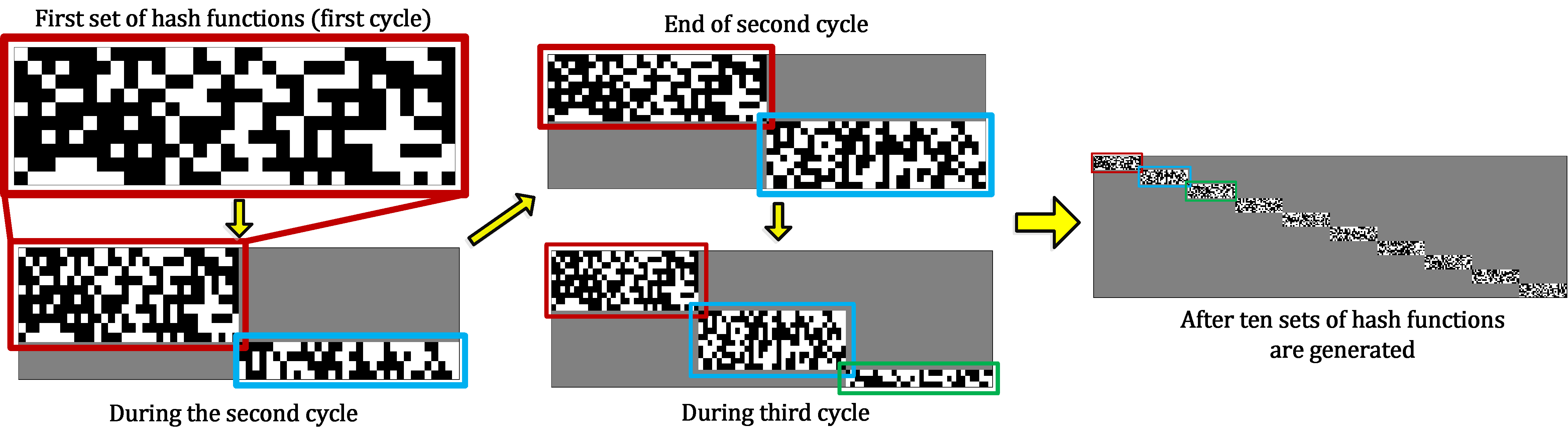}
  \caption{An ECOC Matrix $\Lambda$. Each cycle consists of $\rho$ codewords corresponding to $\rho$ observed labels. At the end of each cycle, $k$ new code words and hash functions are added. As a result, $\Lambda$ expands in both directions. For an incoming point $(\mathbf{x},y)$, only $\rho$ hash functions of the cycle in which the label $y$ was first observed are updated.}
  \label{ecoc_matrix}
\end{figure*}

\subsection{Online/Adaptive Hashing}
Recent studies have tackled the problem of adaptively updating the hash functions with incoming data.

Huang \etal \cite{OKH} recently introduced Online Kernel Hashing, where a stochastic environment is considered with pairs of points arriving sequentially in time. At each step, a number of hash functions are selected based on a Hamming loss measure and an update is done on the hashing parameters via gradient descent. Although a hash function selection strategy exists that reduces binary code re-computations, it does not guarantee a partial selection. For particular parameter selections, the algorithm might require updating all functions at every iteration. 

Yang \etal \cite{SmartHashing} have also proposed a strategy in determining which hash functions to update with the arrival of new data. At each step, a group of hash functions is selected by considering the hash functions' bit assignment consistency or contribution in preserving the similarities as indicated by a similarity matrix. However, this strategy is not online in that previously seen examples are stored and learning is done in batch mode.

Our work is most similar to \cite{OECC} where an online Boosting technique is employed with ECOCs in learning the hash functions. Rather then using a Boosting strategy, we directly minimize a convex upper bound on the Hamming distance between the hash mappings and target codewords. We employ a stochastic gradient technique in which our algorithm has established regret bounds. Also, we introduce a formulation that employs a ternary ECOC matrix; this allows the hash codes to grow over time while, at the same time, reducing visits to previously learned codes stored in an index. As previously stated, it is essential not to re-compute the entire hash table with every update.

\section{Formulation}
\label{formulation}
Let $\mathcal{X}$ and $\mathcal{Y}$ denote the feature and label space, respectively. $\mathbf{z} \triangleq (\mathbf{x},{y})$ is an observation from the joint space $\mathcal{Z} \triangleq \mathcal{X}\times \mathcal{Y}$ where ${y} \in \mathcal{Y}$. We approach hashing as a Hamming embedding. The goal is to find a mapping $\Phi : \mathcal{X} \to \mathcal{H}^b$ where $\mathcal{H}^b$ is the $b-$dimensional Hamming space, so that a particular neighborhood structure is preserved. We specifically focus on preserving the semantic neighborhood as described by label information.  

We utilize a collection of hash functions for this purpose, where a function $f:\mathcal{X} \to \{-1, 1\}$ accounts for the generation of a bit in the binary code. Many types of hash functions are considered in the literature; for simplicity and good performance we utilize linear threshold functions: 
\begin{equation}
\label{hash_function}
f(\mathbf{x})\triangleq\text{sgn} (\mathbf{w}^T\mathbf{x})
\end{equation}
\noindent where $\mathbf{w}$ and $\mathbf{x}$ are in homogeneous form.
\noindent Mapping $\Phi (\mathbf{x}) = [f_i (\mathbf{x}),...,f_b (\mathbf{x})]^T$ then becomes a vector-valued function to be learned.

We will consider using an ECOC formulation for learning $\Phi$. 
The biggest advantage of ECOCs is the error-correcting property, which enables recovery from hash function errors in computing the binary encoding. 
Let $\mathcal{C} \subseteq \mathcal{H}^b$ denote a codebook. Usually the elements of $\mathcal{C}$, \ie~, the codewords, ought to satisfy good error-correcting properties such as having ample bit differences. We will assign each codeword $\mathbf{c} \in \mathcal{C}$ to a label in $\mathcal{Y}$. One intuitive way to preserve the semantic similarity is to minimize the Hamming distance between $\Phi(\mathbf{x})$ and its corresponding codeword $\mathbf{c}_y$. Formally, we would like to find $\Phi$ such that $d_h (\Phi(\mathbf{x}), \mathbf{c}_y)$ is minimized where $d_h(\mathbf{a},\mathbf{b})$ is the Hamming loss/distance between $\mathbf{a}$ and $\mathbf{b}$. Hence, the objective can be formulated as:
\begin{equation}
\label{main_objective}
J(\Phi) \triangleq \mathbb{E}_{\mathcal{Z}} \left[d_h(\Phi(x),\mathbf{c}_y)\right] \triangleq \int_{\mathcal{Z}} d_h(\Phi(x),\mathbf{c}_y) dP(\mathbf{z}).
\end{equation}

We would like to solve Eq.~\ref{main_objective} in an online manner. This allows the computations to grow linearly with data input size, which is crucial given large-scale datasets.  

The Hamming distance is  defined as $d_h( \Phi(\mathbf{x}), \mathbf{c}_y) = \sum_t [\![f_t(\mathbf{x}) \neq c_{yt}]\!]$  where both $d_h$ and the functions $f_t$ are non-differentiable. Fortunately, we can relax $f_t$ by dropping the $sgn$ function in Eq.~\ref{hash_function} and derive an upper bound 
on the Hamming loss. Note that $d_h( \Phi(\mathbf{x}), \mathbf{c}_y) = \sum_t [\![f_t(\mathbf{x}) \neq c_{yt}]\!] \leq \sum_t l(-c_{yt} \mathbf{w}_t^T \mathbf{x})$ with a suitably selected convex margin-based function $l$. Thus, by substituting this surrogate function into Eq.~\ref{main_objective}, $J(\Theta) (> J(\Phi))$ becomes a convex where $\Theta^T = [\mathbf{w}_1^T,...,\mathbf{w}_T^T]$. Therefore, we can directly minimize this upper bound using stochastic gradient descent:
\begin{equation}
\label{sgd_update_rule}
\Theta^{t+1} \leftarrow \Theta^t - \eta_t \nabla_{\Theta} J(\Theta)
\end{equation}
\noindent where $\eta_t > 0$. Zinkevich \cite{Zinkevich03onlineconvex} showed that for SGD, requiring $J(\Theta)$ to be a differentiable Lipschitz-function is sufficient for obtaining a regret bound that diminishes as $\mathcal{O}({1}/{\sqrt{T}})$, where $T$ is the iteration number. 

Expanding and diversifying datasets are common in many practical problems. Hence, it is crucial that the mapping accommodates these variations. In our online setting, an incoming point may be associated with a previously observed class label or may even be associated with a new 
class label. The online framework we employ allows the hash functions to be adapted with streaming data. If a new label is observed, we assign a unique codeword from $\mathcal{C}$ to the label and proceed on with the minimization.

\medskip
\noindent \textbf{Reducing Hash Table Updates.} As discussed, the concept of \textit{adaptive} hash functions is extremely appealing; though, it could require that all previously computed hash codes must be updated with every change of $\Phi$. Assume $N$ $b$-length hash codes are already stored in an index. 
This requires updating $Nb$ bit entries in the indexing structure. After $T$ iterations a total of $\mathcal{O}(NbT)$ computations must be carried out to keep the index up to date. Although, hash function evaluation is usually fast, if $N$ is large and the index is partially stored on a disk, these computations may require disk access and become extremely inefficient. 
Hence, a solution that reduces hash bit entry updates in indexing is needed. 

To address this issue we utilize ternary codewords $\mathbf{c}_y \in \{-1, 0, 1\}^b$, where $0$ denotes an \textit{inactive} bit. As used in classification problems \cite{Escalera}, the inactive bit allows us to avoid any changes to the corresponding (inactive) generator hash function. 
The Hamming distance $d_h$ now becomes $d_h (\Phi(x), \mathbf{c}_y) = \sum_t [\![f_t(\mathbf{x}) \neq c_{yt}]\!]|c_{yt}|\leq \sum_t l(-c_{yt} \mathbf{w}_t^T \mathbf{x})|c_{yt}|$. Hence, as the inactive bits do not contribute to the Hamming distance, their partial derivative $\nabla_{\mathbf{w}_t} J$ in Eq.~\ref{sgd_update_rule} is $\mathbf{0}$. 

Our framework is online and the number of labels is not known {\em a priori}. Consequently, rather than fixing the code length parameter $b$, we systematically grow the binary codes with incoming data. Please observe Fig.~\ref{ecoc_matrix}. Assume we store codewords $\mathbf{c}_y$ corresponding to observed labels as rows in an ECOC matrix $\Lambda$. A \textit{cycle} in our method consists of observing $\rho$ new labels; for example, the first set of $\rho$ labels observed during learning constitute the first cycle. At the end of each cycle, we alter the matrix $\Lambda$ by appending $k$ {\it inactive} columns to the right. Let $m$ denote the number of cycles so far. The next incoming newly observed label is assigned the ternary code $\mathbf{c}_y = [0\cdots0~\overline{\mathbf{c}}_y]$ with $(m-1)k$ inactive bits and a $k$-length binary codeword $\overline{\mathbf{c}}_y \in \mathcal{C}$.  Therefore, $\Lambda$ grows in both directions with streaming data. We summarize our algorithm in Alg.~\ref{alg:algorithm1}.

\IncMargin{1em}
\begin{algorithm}[t!]
\caption{}
\label{alg:algorithm1}
\small
\LinesNumbered\DontPrintSemicolon
\SetKwData{Left}{left}\SetKwData{This}{this}\SetKwData{Up}{up}\SetKwData{Ind}{ind}
\SetKwFunction{Union}{Union}\SetKwFunction{FindCompress}{FindCompress}
\SetKwInOut{Input}{input}\SetKwInOut{Output}{output}
\Input{Codebook $\mathcal{C}, \rho, k, ~\eta_t$, streaming inputs $(\mathbf{x}^t,y^t)$, initiliazation procedure for a set of hash functions \texttt{init()}, procedure \texttt{find($\Lambda, y$)} to obtain the codeword of $y$ from code matrix $\Lambda$,}
$\Theta =$ \texttt{init}$(\mathbf{w}_1,\ldots,\mathbf{w}_k), m=1, n=1$\;
\BlankLine
\For{$t\leftarrow 1,..., T$}{
$flag \leftarrow 0$\;
\If(\tcp*[h]{A new label}){${y}^t \notin \mathcal{Y} $}{
 $flag \leftarrow 1$\;
\eIf(\tcp*[h]{A new cycle}){$n > \rho$}{
$\Lambda \leftarrow [\Lambda~\mathbf{0}^{\rho \times k}]$\;
$\Theta = [\Theta $ \texttt{init}$(\mathbf{w}_{mk+1},...,\mathbf{w}_{(mk+k})]$\;
$ m \leftarrow m + 1, n \leftarrow 1$\;
}{
$n \leftarrow n + 1$\;
}
$\overline{\mathbf{c}}_y \leftarrow$ random codeword from $\mathcal{C}$\;
$\mathcal{C} \leftarrow \mathcal{C}\backslash \overline{\mathbf{c}}_y$\;
$\mathbf{c}_y \leftarrow [\overbrace{0,..,0}^{(m-1)k}~\overline{\mathbf{c}}_y]$ \tcc*{Ternary code assigned to label $y$}\;
$\Lambda \leftarrow \left[ \begin{matrix}
   \Lambda\\
   \mathbf{c}_y  \\
\end{matrix} \right]
 $
}
\lIf{$flag = 0$}{$\mathbf{c}_y \leftarrow$ \texttt{find($\Lambda, y$);}}
\;
\;
$\Theta^{t+1} \leftarrow \Theta^t - \eta_t \nabla_{\Theta} J(\Theta;\mathbf{c}_y)$ \tcp*{Update}\;

}
\end{algorithm}\DecMargin{1em}

\subsection{Parameter Selection}
\label{parameter_selection}

Alg.~\ref{alg:algorithm1} requires user-specified choices for $\mathcal{C}, \rho$ and $k$. Though the performance does not critically depend on delicate parameter choices, we still discuss selection heuristics. 

Central to the method is the codebook $\mathcal{C}$, where its elements $\overline{\mathbf{c}}_y \in \{-1,1\}^k$ are used to construct target hash codes for the mappings. Though these binary codes can be computed on the fly with demand, it is more appropriate to construct them offline to reduce online computations. These codes ought to have good row and column separation properties. We consider a randomly generated codebook, which is simple to construct and shown to perform better compared to other heuristics~\cite{Rep1}. 

Though we assume no prior information for the label space, let $L$ denote the total number of labels we anticipate. 
During learning, the ternary codewords are constructed from codes sampled from the binary codebook $\mathcal{C}$. Since the inactive bits in the ternary codewords do not contribute to the Hamming distance, the binary codes $\overline{\mathbf{c}}_y \in \mathcal{C}$ instead must have ample bit differences. To ensure this property, codebook $\mathcal{C}$ must have a sufficiently large number of codes. Specifically, $2^k \gg  L$ must be satisfied where $k$ is the length of $\overline{\mathbf{c}}_y$. This is a mild requirement given the common choices of $k$ ($k=32, 64, \dots$) which work well in practice. 

Another situation to avoid is $\rho < \log k$, as this can lead to identical columns (bipartitions) in a cycle. To examine why we need $\rho \geq \log k$, consider the codewords assigned to $\rho$ labels in a particular cycle. Ignoring the inactive bits, notice that the columns denote bipartitions of the data based on labels in which the hash functions can be considered as dichotomizers. Eq.~\ref{main_objective} can be reorganized by interchanging the integral and sum, yielding $J(\Theta)=\sum_t \int_{\mathcal{Z}} l(-c_{yt} \mathbf{w}_t^T \mathbf{x})dP(\mathbf{z}) $. Hence, with the upper bound, minimizing the Hamming loss between the target hash codes $\mathbf{c}_y$ and the mapping $\Phi(\mathbf{x})$ can be viewed as minimizing the sum of expected losses or the binary classification errors. If identical bipartitions exist then the learned dichomotizers will be highly correlated as the classification error of each dichomotizer will independently be minimized based on these same bipartitions (data splits). 

In general, $k$ must be selected so that the inequality $2^k \gg L \geq \rho \geq \log k$ holds. For randomly generated $\rho$ codes of $k$-length, the probability that all bipartitions are unique is $\frac{2^{\rho}!}{(2^{\rho}-k)!2^{\rho k}}$. Thus, a good choice for $\rho$ might be $4 \log k$ where this probability is $\geq 0.9$ for common choices of $k$.

\begin{figure}[t]
  \centering
      \includegraphics[width=0.44\textwidth]{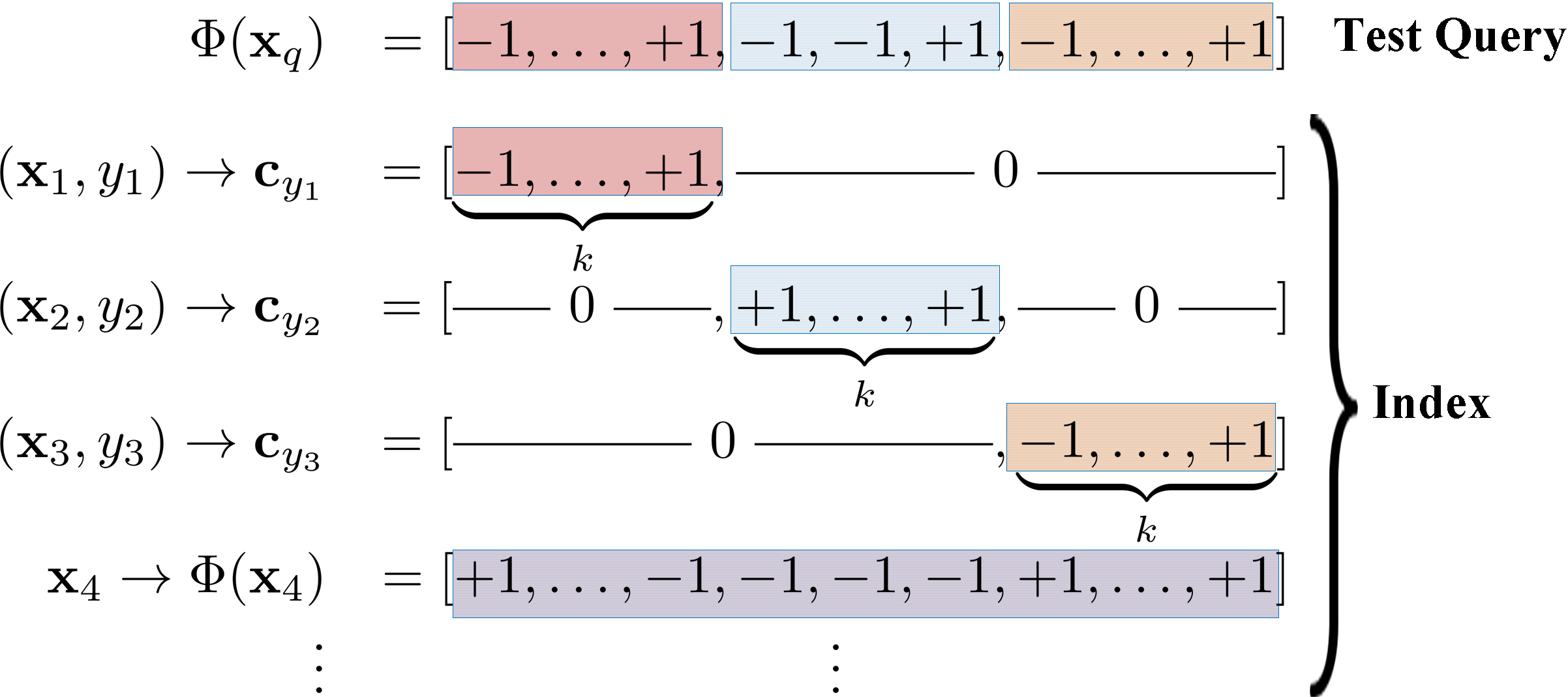}
  \caption{A point $\mathbf{x}$  may be indexed by either using its corresponding codeword $\mathbf{c}_y$ (if label information exists) or by using the output of the mapping, \ie~, $\Phi (\mathbf{x})$. As seen, for ternary codewords the inactive bits do not contribute to the Hamming distance (only highlighted areas are compared). }
  \label{index}
\end{figure}

\subsection{Populating Hash Table and Retrieval}
\label{decoding}

After learning the mapping $\Phi$, the hash table can be populated by either indexing a point $\mathbf{x}$ using its corresponding codeword $\mathbf{c}_y$ in $\Lambda$ or by using the output of the mapping, \ie~, $\Phi (\mathbf{x})$ (see Fig.~\ref{index}). Given a test query $\mathbf{x}_q$, $\Phi(\mathbf{x}_q)$ is computed and instances are ranked based on the Hamming distances between the binary codes. Note that $\Phi(\mathbf{x}) \in \{-1,1\}^b$ while the hash codes of the points are ternary, as described above. Even with ternary codes, the Hamming distance computation can be carried out with the same efficiency by masking out the inactive bit positions. 

If the point to be indexed has label information $y$, then using the corresponding codeword as its binary code is shown to perform better \cite{ECC}. Instances can be ranked by merely computing the Hamming distance between the test query code $\Phi(\mathbf{x}_q)$ and the codewords. Since the number of codewords is usually much smaller than the number of instances, this is much more efficient and the ECC error-correcting property also allows a number of hash function errors to be compensated for during this retrieval. Moreover, since the hash table is populated via codewords $\mathbf{c}$, an update to previously stored entries is not required when the mapping $\Phi$ is changed. This is an important property for an adaptive hashing technique since updating the index for each change in the hash mapping might be computationally infeasible for large structures. 

However, label information may not always be available. In such a case, we can use $\Phi$ to compute the binary code of an instance. However, such codes must be accommodated to the changes to $\Phi$; thus, reducing this need to update the hash table becomes essential. We do this by allocating sets of hash functions to sets of classes with the usage of ternary codewords as explained Sec. \ref{formulation} and Alg. \ref{alg:algorithm1}.

\section{Experiments}
\label{experiments}
In this section we evaluate and compare our method against state-of-the-art techniques on two standard image retrieval benchmarks. For comparison, we consider Binary Reconstructive Embeddings (BRE) \cite{BRE}, Minimal Loss Hashing (MLH) \cite{MLH}, Supervised Hashing with Kernels (SHK) \cite{SHK}, Fast Supervised Hashing (FastHashing) \cite{FastHash}, Supervised Hashing with Error Correcting Codes (ShECC) \cite{ECC}, Smart Hashing (SmartHash) \cite{SmartHashing}, Hashing by Deep Learning \cite{aaai_hash,Lin_2015_CVPR_Workshops,Lai_2015_CVPR}, Online Kernel Hashing (OKH) \cite{OKH} and Online Supervised Hashing (OSH) \cite{OECC}. These methods have outperformed earlier methods such as \cite{LSH, SH, SSH, SSC, STH, LDAHASH}. Similar to our work, OKH and OSH are online hashing methods. 

In evaluating retrieval performance we consider the following scheme. Given a query $\mathbf{x}_q$, $\Phi(\mathbf{x}_q)$ is computed and instances are ranked according to their Hamming distances to the query. This linear search is extremely efficient in Hamming space. To quantify our results, we report the Mean Average Precision (mAP). We consider binary codes of length 8 to 64. 

We further compare our method with OKH, OSH and SmartHash to analyze the number of indexed binary code updates during a learning phase. We specifically report mAP values \vs the number of bit updates with different choices of parameters.

\subsection{Experimental Setup}
\label{exp_setup}
We conduct evaluation on two datasets: Cifar-10 and Sun397.
We follow common guidelines in constructing our training and test sets \cite{SHK, SSH}. Unless otherwise stated, the training and test sets are constructed with an equal number of samples from all labels. The rest of the data is used to populate the hash table. If label information is provided, then we index a point with the codeword corresponding to its label. If a label is not provided, $\Phi$ is used for indexing.

Given a test query, the goal is to retrieve (or rank highly) instances associated with the same label. We use the same data splits for all batch-learning methods. For online techniques, the experiments are repeated five times with different orderings of the incoming data and the average performance values are reported. Different types of hash functions such as linear and kernel SVMs, decision trees, etc.\  are available for certain methods such as ShECC and FastHash. We simply select the best performing type of function. 

\begin{figure}[t]
  \centering
      \includegraphics[width=0.3\textwidth]{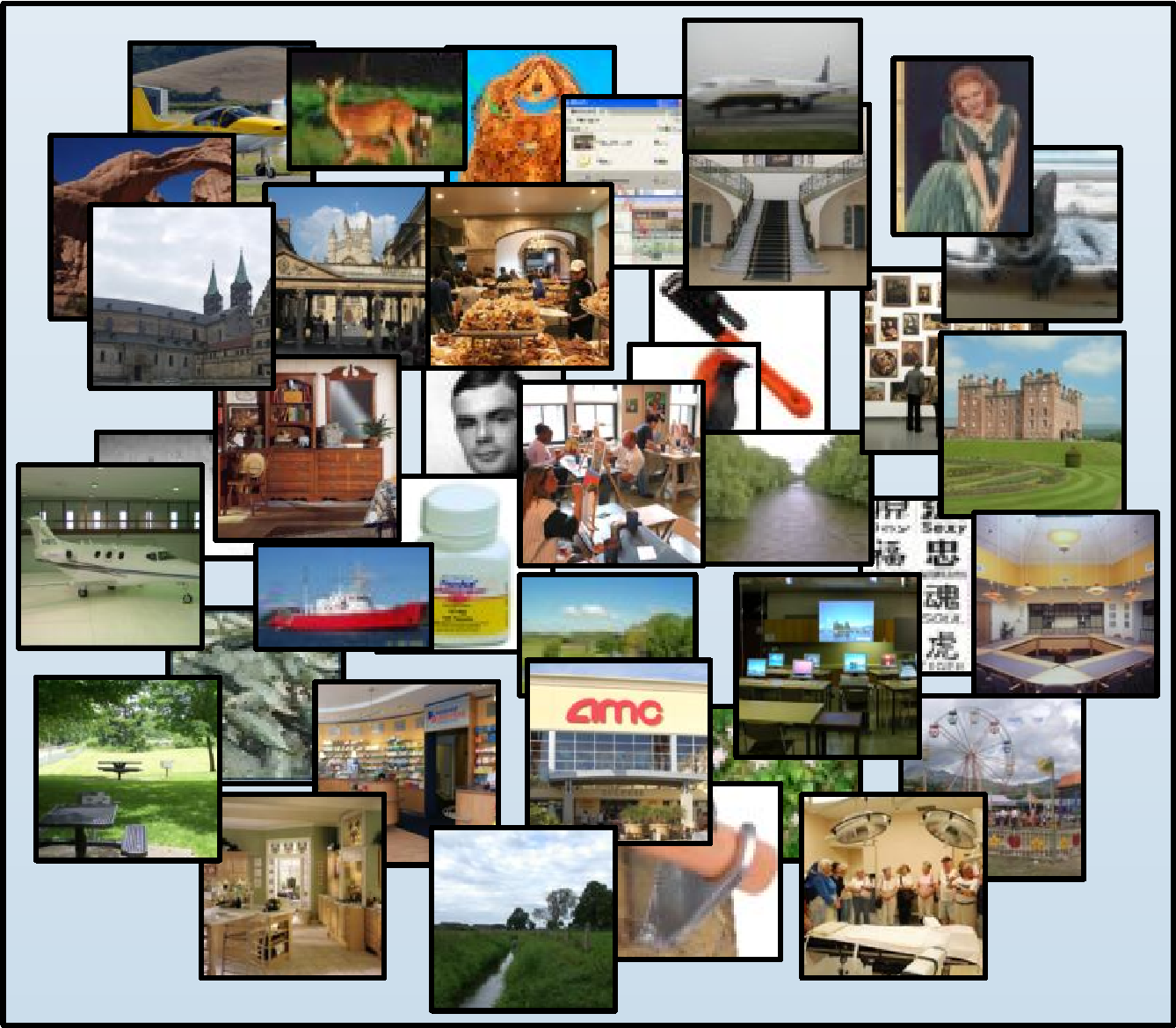}
  \caption{Example images from the Cifar-10 and Sun397 datasets.}
  \label{images}
\end{figure}

In Sec.~\ref{ret_perf}, when evaluating retrieval performance, we set $\rho$ sufficiently large so that only a single cycle occurs in Alg.~\ref{alg:algorithm1}. The value of $k$ then denotes the bit-length (8 to 64). In Sec.~\ref{up_perf}, we compare the adaptive methods based on the number of bit updates applied to already indexed data during learning. In so doing, we allow multiple cycles in Alg.~\ref{alg:algorithm1} and report retrieval performance by using different parameter values. Notably, the step size $\eta_t$ is set to $1$ in all experiments. 

We use the Gist descriptor and CNN features as image representations. The CNN features are obtained from the $fc7$ layer of a VGG16 \cite{simonyan2014very} Convolutional Neural Network pre-trained on ImageNet \cite{deng2009imagenet}. Note that we \textit{do not} fine-tune the network to the target datasets Cifar-10 and Sun397. Finally, some of the compared methods mean-center and unit normalize the input features; therefore, such preprocessing of features is applied to put all compared methods on equal footing.  

Regarding Smart Hashing, this procedure is not a hashing method but a strategy to update the hash functions in a selective manner. This strategy can be used in conjunction with suitable hashing methods to reduce computational time. Similar to their work, we apply this framework to SHK with the default parameters as noted by the authors. We consider the similarity based selection scheme which is reported to be better performing. The number of hash functions updated with each incoming set of points is set to $\lceil{\log{b}}\rceil$ where $b$ is the total length of the embedded binary codes. The incoming data set size ($\triangleq q$) is chosen to be 100 while we also begin initially with 100 samples. 

\medskip
\noindent {\bf Cifar-10} contains 60K images from 10 different categories. Example categories include: airplane, deer, ship, truck, etc. We split the dataset into 59K training and 1K testing images. 2K images are further sampled from the training set for learning the hash functions.  

\medskip
\noindent {\bf Sun397} contains over 108K images from 397 scene categories. There are at least 100 images per category. We split the dataset into two: a training and test set with 104K and 3970 images, respectively. The 3970 training images are also used to learn the hash functions. 

\begin{table}[t!]

  \centering\scriptsize
  
{
\hfill{}
    \begin{tabular}{c|rrrr|c}
    \hline
\hline
    \textbf{Method} &  \multicolumn{4}{|c|}{\begin{tabular}{@{}c@{}}\textbf{Mean Average Precision }  \\ \textbf{Random}  0.1  \end{tabular}}   & \begin{tabular}{@{}c@{}}\textbf{Training Time} \\ \textbf{(seconds)} \end{tabular}\\
    \hline \hline
        \# of bits  & 8     & 16    & 32    & 64    & 32  \\
\hline
    BRE \cite{BRE}  & 0.16  & 0.17  & 0.16  & 0.17  & 263 \\
    MLH  \cite{MLH} & 0.16  & 0.17  & 0.18  & 0.19  & 224\\
    SHK  \cite{SHK} & 0.23  & 0.26  & 0.30  & 0.31  & 1115 \\
    SmartHash \cite{SmartHashing} & 0.24  & 0.27  & 0.28  & 0.30  & 604 \\
    FastHash \cite{FastHash}& 0.27  & 0.33  & 0.34  & 0.37  & 194 \\
    ShECC \cite{ECC} & \textbf{0.41}  &\textbf{0.50}  &\textbf{0.57}  & \textbf{0.58}  & 1075 \\
\hline
    OKH  \cite{OKH} & 0.12  & 0.13  & 0.14  & 0.15  & 8.7 \\
    OSH  \cite{OECC} & 0.39  & 0.44  & 0.48  & 0.53  & 5.5 \\
   
    Ours  & \textbf{0.41}  & {\color{red}0.46}  & {\color{red}0.50}  & {\color{red}0.55}  & \textbf{2.3} \\
        \hline 
        \hline
          ShECC (CNN) & 0.71  & 0.76  & 0.81  & 0.83  & 1608 \\
              OSH (CNN) & 0.62  & 0.69  & 0.75  & 0.76  & 13 \\
        Ours (CNN, $\Phi$)  & 0.47  &  0.51 & 0.60 &  0.63 & 5.2\\
            Ours (CNN) & 0.70 & 0.76  & 0.80 & 0.83 & 5.2 \\
\hline
     \hline
    \end{tabular}
}
\hfill{}
\caption{Mean Average Precision and training time @ 32 bits for Cifar-10. \textbf{Bold} denotes the best performing method while {\color{red}{red}} denotes the best online technique.}
  \label{CifarMapTable}
\end{table}

\begin{table}

  \centering\scriptsize
{
\hfill{}
    \begin{tabular}{c|rrrr|c}
    \hline
\hline
    \textbf{Method} &  \multicolumn{4}{|c|}{\begin{tabular}{@{}c@{}}\textbf{Mean Average Precision ($\times 10^{-1}$)}  \\ \textbf{Random}  0.026  \end{tabular}}   & \begin{tabular}{@{}c@{}}\textbf{Training Time} \\ \textbf{(seconds)} \end{tabular}\\
    \hline \hline
        \# of bits  & 8     & 16    & 32    & 64    & 32  \\
\hline
    BRE  \cite{BRE} & 0.052 & 0.059 & 0.062 & 0.074 & 301 \\
    MLH  \cite{MLH} & 0.049 & 0.065 & 0.080 & 0.079 & 293 \\
    SHK  \cite{SHK} & 0.058 & 0.057 & 0.054 & 0.051 & 860 \\
    SmartHash \cite{SmartHashing} & 0.053 & 0.053 & 0.053 & 0.053 & 2800 \\
    FastHash \cite{FastHash} & 0.044 & 0.050 & 0.059 & 0.072 & 2400 \\
    ShECC \cite{ECC} & 0.086 & \textbf{0.128} & {0.146} & 0.191 & 11262 \\
\hline
    OKH  \cite{OKH} & 0.034 & 0.043 & 0.053 & 0.062 & 19 \\
    OSH \cite{OECC}  & 0.096 & 0.110 & 0.140 & 0.190 & 9.2 \\
    Ours  & \textbf{0.100} & {\color{red}0.120} & {\textbf{0.160}} & \textbf{0.220} & \textbf{5.2} \\
\hline
\hline
    ShECC (CNN)        & 0.140 & 0.210 &  0.320 & 0.570 & 14574 \\
    OSH (CNN)          & 0.077 & 0.079 & 0.089 & 0.541 & 27 \\
    Ours (CNN, $\Phi$) & 0.062 & 0.092 & 0.164 & 0.301 & 6.9 \\
    Ours (CNN)         & 0.166 & 0.375 & 0.567 & 0.790 & 7.2 \\
\hline
     \hline
    \end{tabular}
}
\hfill{}
\caption{Mean Average Precision and training time @ 32 bits for Sun397. \textbf{Bold} denotes the best performing method while {\color{red}{red}} denotes the best online technique.}
  \label{SunMapTable}
\end{table}

\subsection{Retrieval Performance} 
\label{ret_perf}

Table~\ref{CifarMapTable} shows results for the Cifar-10 dataset. 
There are three main groups of results presented in this table.  The first group corresponds to results using batch methods trained with the Gist descriptor. The second group comprises online methods trained with Gist. The last group of rows in the table shows the top competing batch and online method, in which CNN features are used as the image representations.

We observe ShECC performs best for all length binary codes. Our method is the overall runner-up, while being the top performing online solution. Our method surpasses every other batch-learning technique with up to four orders-of-magnitude speedup (\textbf{1073} \vs \textbf{2.3} seconds of learning time for ShECC and our method, respectively). We outperform the Online Kernel Hashing solution (OKH) with significant Mean Average Performance boosts, \eg, \textbf{0.12} \vs \textbf{0.41} $(\mathbf{+242\%})$ for 8 bits and \textbf{0.15} \vs \textbf{0.55} $(\mathbf{+267\%})$ for 64 bit codes. Compared to OSH, our technique also demonstrates both retrieval accuracy and computational time improvements. 

With the CNN features we observe a significant boost in performance for all the techniques. Similarly to previous results, our method is the best performing online method and the overall runner up technique showing comparable results to ShECC but being much faster.  

Table~\ref{SunMapTable} shows results for the Sun397 dataset. Our method is again either the overall best performing technique or the runner-up. With only 8 bits our retrieval performance is more accurate compared to all other techniques excluding ShECC. We achieve these results with drastic learning time improvements (only \textbf{5.2} seconds). One interesting observation is: SmartHash is slower to learn its hash functions compared to SHK. This is due to the reason that, although updating a subset of hash functions is faster (only $\lceil \log(b) \rceil$ among $b$ hash functions are updated), the previously observed data is not discarded with the new incoming set of points. At each step, $\lceil \log(b) \rceil$ functions are updated but in a batch-learning procedure for this larger set of points, eventually yielding slow computational times. 

Similar to the Cifar-10 benchmark, we see a boost in performance when CNN features are used as image representations. Our method improves over the competing online and batch method OSH and ShECC (\textbf{0.79} \vs \textbf{0.54} \& \textbf{0.57} $(\mathbf{ + 46 \%})$  for 64-bit codes, \textbf{0.5} \vs \textbf{0.08} \& \textbf{0.32} $(\mathbf{ + 66 \%})$ for 32-bit codes).

We also provide results when $\Phi$ is used for indexing the data. This might be the case when the label of an instance is missing or not available. The retrieval scores are not as high as when codewords are used; however, the results show overall good performance. Utilizing the codewords enables error correcting by mapping instances to codes that have good separation. Thus, in retrieval (where the label for the query is unavailable), a number of hash function errors can be compensated for. However, when $\Phi$ is used for indexing, the Hamming distance between binary codes of points from distinct classes can be small, degrading the retrieval performance. For these reasons, whenever a label exists for an instance, it is much better to use the corresponding codeword as its binary code for indexing.  

\subsubsection{Comparison with Deep Learning Methods} 
We also compare our method against recently introduced deep learning based hashing approaches \cite{aaai_hash,Lin_2015_CVPR_Workshops,Lai_2015_CVPR}, using the Cifar-10 benchmark. These methods simultaneously learn features for the target dataset and hash functions. As the performance boost in such methods is also due to feature learning, we use CNN features in our method for fair comparison. These features are obtained from the $fc7$ layer of a VGG16 Convolutional Neural Network pre-trained on ImageNet. We note that the features used in our method \textit{are not} based on a network that is fine-tuned on our target datasets. We also use the same evaluation setup as noted in the respective papers. Specifically, 5K and 50K training images are used when comparing against \cite{aaai_hash,Lai_2015_CVPR} and \cite{Lin_2015_CVPR_Workshops}, respectively.  Also, such deep learning methods often train multiple epochs over the data, whereas we merely process the data only once. 

In the experiments with the Cifar-10 benchmark, we observe a mAP value of \textbf{0.803} for our method when 5K images are used for training. Please compare this to \textbf{0.581} $\mathbf{(+ 37 \%)}$ and \textbf{0.532} $\mathbf{(+ 51 \%)}$ for \cite{aaai_hash} and \cite{Lai_2015_CVPR}, respectively. When trained with 50K images we obtain a mAP value of \textbf{0.87}. In \cite{Lin_2015_CVPR_Workshops}, authors report a \textbf{0.89} mAP value. However, notice that our CNN features are not fine-tuned. Also, it takes $\sim$\textbf{12} and $\sim$\textbf{150} seconds for training with 5K and 50K points, respectively, on a standard PC CPU to get better or at least comparable results; opposed to training for a few hours on a GPU for the deep learning methods. 

\subsection{Update Performance}
\label{up_perf}
As stated above, adaptive hashing is appealing; but, it may require revisiting and recomputing the hash codes for previously indexed items as the hash functions are updated. In this section, we report the number of bit updates in an index during a learning phase for SmartHash, OKH, OSH, and our method. Learning is done with the same setup as described in Sec.~\ref{exp_setup}. We first provide a complexity analysis and then interpret evaluation results.

\medskip
\noindent \textbf{Complexity Analysis.}
SmartHash and OKH update a subset of hash functions at each step. The corresponding bits of all codes in an index are subsequently updated. Assume we have $N$ binary codes in our index. SmartHash relearns $z$ hash functions in each iteration; hence, it requires $z$ bit entry updates for a single binary code. Assume $\hat{T}$ is the iteration number, then SmartHash carries out $\mathcal{O}(zN\hat{T})$ computations to keep the index up to date. For OSH this number is $\mathcal{O}(NbT)$ where $b$ and $T$ are the length of the binary code and the iteration number, respectively.\footnote{Note that $\hat{T}$ and $T$ are different. Given the setup in Sec.~\ref{exp_setup}, if $M$ is the number of training points to learn the hash functions, then $\hat{T} = M/q$ and $T = M$.} For OKH, the number of bit updates depends solely on algorithmic parameters. Hence, we report runtime measurements with only the cross-validated parameters.

For our method, when $\rho$ is set sufficiently large so that only a single cycle occurs in Alg.~\ref{alg:algorithm1}, the complexity of keeping the index up to date after $T$ iterations is the same as OSH, namely $\mathcal{O}(NbT)$. For analysis purposes, we forgo the no prior assumption on the number of labels to examine the case where multiple cycles might occur. Assume there are $L$ labels in total and $L \geq \rho$. Also assume we are in a state where we observed all labels and the incoming points are uniformly sampled from all such $L$ labels. In such a case, updating the index requires $\mathcal{O}(\frac{\rho NbT}{L})$ computations where the binary code length is now $b = \lceil \frac{kL}{\rho} \rceil$. Hence, a tradeoff exists between reducing the hash bit updates and the total code length. Fortunately, as seen in Fig.~\ref{ecoc_matrix}, $\Lambda$ is sparse in that most of the entries of the codewords are inactive bits. With a small overhead of computation we can exploit this sparsity and we need not store the entire codeword in memory during Hamming distance computations. 
\begin{figure*}[t!]
        \centering
        \begin{subfigure}[b]{0.36\textwidth}
                \includegraphics[width=\textwidth]{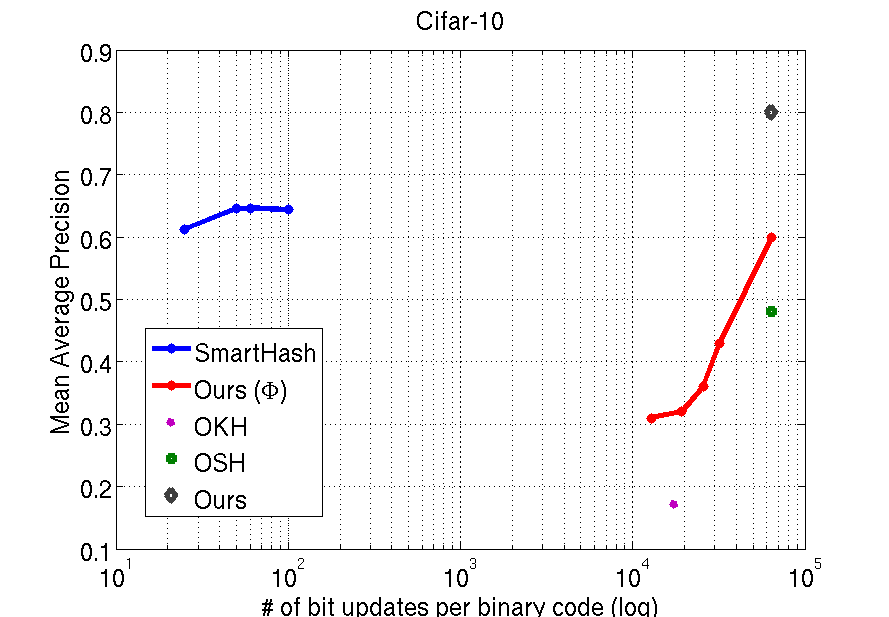}
        \end{subfigure}
        ~ 
        \begin{subfigure}[b]{0.36\textwidth}
                \includegraphics[width=\textwidth]{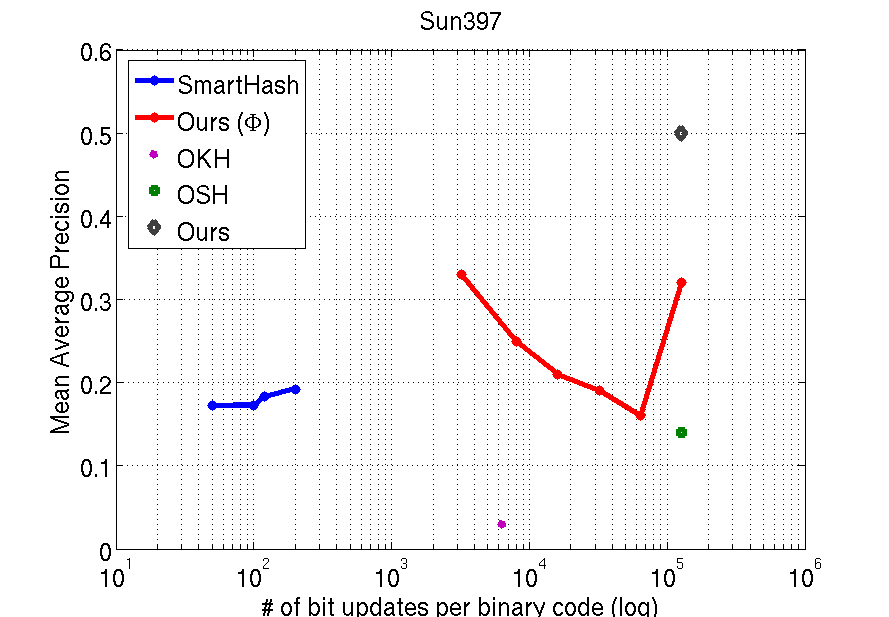}
        \end{subfigure}
        ~ 
        \caption{Mean Average Precision with respect to number of bit updates occurrences per indexed point during learning.}\label{fig:hash_update}
\end{figure*}

\begin{figure}[t!]
  \centering
      \includegraphics[width=0.3\textwidth]{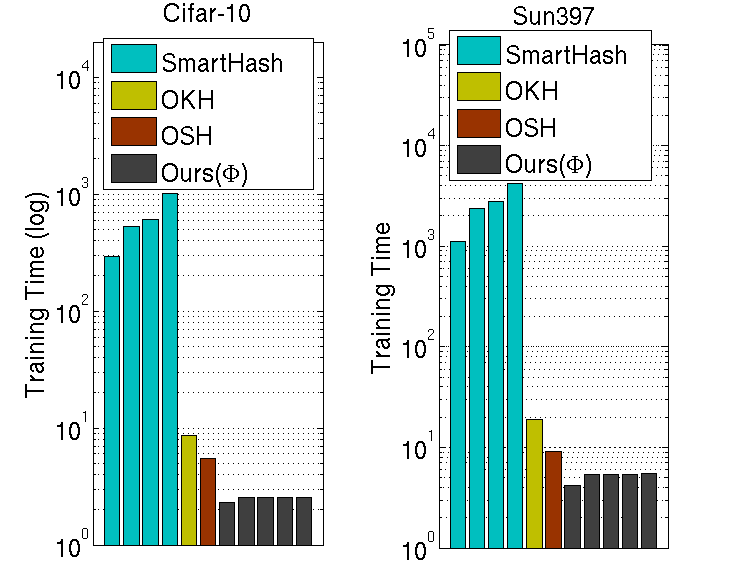}
        \caption{Training time for \textit{SmartHash, OKH, OSH} and \textit{Ours ($\Phi$)}. The columns of \textit{SmartHash} and \textit{Ours ($\Phi$)} correspond to the parameter choices in their given order as noted in Sec.~\ref{up_perf}. Training time for \textit{Ours} is similar to \textit{Ours ($\Phi$)}.}\label{fig:hash_update_tt}
\end{figure}

\medskip
\noindent \textbf{Runtime Analysis.} Fig.~\ref{fig:hash_update} shows the retrieval performance with respect to the number of bit updates per indexed point. Following \cite{SmartHashing}, we set the parameters in the SmartHash method to $(z,q) = \{(5,400),(5,200),(3,100),(5,100)\}$. The code length is set to $32$. For our method, we set the number of hash functions in each cycle to be $k=32$ where each cycle consists of $\rho = c\log(k)$ codewords. Specifically, $c = \{40,20,10,5,2\}$ for the Sun397 benchmark and $\rho = \{5,4,3,2\}$ for Cifar-10. Finally, for OKH and OSH, we report the mAP value and number of bit updates per binary code for a single set of cross-validated parameters. 

Fig.~\ref{fig:hash_update} shows the number of bit updates per indexed data point during learning with respect to Mean Average Precision (mAP). While we observe SmartHash relearns hash codes less frequently, the retrieval performance is inferior to our method. Also as shown in Fig.~\ref{fig:hash_update_tt} the computation time for learning in SmartHash is orders-of-magnitude greater than with our method. Regarding OKH, this method requires fewer updates but its mAP is inferior to our method. 

When the mapping $\Phi$ is used for indexing, the hash table must be updated to accomodate the changes in the hash mapping. With Alg.~\ref{alg:algorithm1} we can control the number of bit updates that occur during learning. For Cifar-10, we observe a tradeoff between the number of bit updates \vs mAP performance. This is different for the Sun397 benchmark where reduction in bit updates does not necessarily mean lower retrieval performance. For datasets with a relatively small number of classes (\eg, Cifar-10) allocating different sets of hash functions to separate classes will cause the retrieval performance to drop due to such sets of hash functions only being discriminatory with respect to individual classes. However, for datasets with larger numbers of classes, such an allocation might have a positive effect since it may be hard to bipartition data in such an intricate feature space with lots of classes \cite{Rep1}. The parameter choices should be selected to suit the specific application.

Overall, we observe that our method either performs best in term of retrieval accuracy and number of updates in the index with respect to OSH and OKH. Compared to SmartHash, our method shows improved retrieval perfomance but with higher number of hash table updates, but the training time is orders-of-magnitude faster as shown in Fig. \ref{fig:hash_update_tt}. Also, SmartHash is not an online method requiring to maintain previously seen data instances.

\section{Conclusion}
\label{conclusion}

We introduce an online hashing method that achieves state-of-the-art retrieval performance. As observed in the experiments, our method is orders-of-magnitude faster compared to batch solutions and significantly faster than competing online hashing methods. 
Online hashing methods are important due to their ability to adapt to variations in datasets. Equally important, online hashing methods offer superior time complexity \vs batch methods for learning with respect to dataset input size. 

{\footnotesize
\bibliographystyle{ieee}
\bibliography{egbib}
}
\end{document}